\pdfoutput=1
\documentclass[11pt]{article}
\usepackage[utf8]{inputenc}
\usepackage{algorithm}
\usepackage{algpseudocode}
\usepackage{array}
\usepackage{xcolor}
\usepackage{svg}
\usepackage{changepage} 
\newcommand{\argmax}{\operatornamewithlimits{argmax}}
\usepackage{amsmath}
\usepackage{acl}
\usepackage{times}
\usepackage{latexsym}
\usepackage{graphicx}
\usepackage{booktabs}
\usepackage{algorithm}
\usepackage{algpseudocode}
\usepackage{array}
\usepackage[export]{adjustbox}
\usepackage{tabularx} 
\usepackage{adjustbox}
\usepackage[T1]{fontenc}

\usepackage[utf8]{inputenc}

\usepackage{microtype}

\usepackage{enumitem}
\usepackage{inconsolata}
\usepackage[labelfont=bf]{caption}
\usepackage{color}

\title{Precision Empowers, Excess Distracts:\\ Visual Question Answering With Dynamically Infused Knowledge In Language Models}

\author{Manas Jhalani, Annervaz K M \textbf{and} Pushpak Bhattacharyya \\
  Computer Science and Engineering, IIT Bombay\\Indian Institute of Science, Bangalore\\
  \{\texttt{manasj, pb}\}@cse.iitb.ac.in\\annervaz@iisc.ac.in}

\begin{document}
\maketitle
\begin{abstract}
In the realm of multimodal tasks, Visual Question Answering (VQA) plays a crucial role by addressing natural language questions grounded in visual content. Knowledge-Based Visual Question Answering (KBVQA) advances this concept by adding external knowledge along with images to respond to questions. We introduce an approach for KBVQA, augmenting the existing vision-language transformer encoder-decoder (OFA) model~\citep{wang2022ofa}. Our main contribution involves enhancing questions by incorporating relevant external knowledge extracted from knowledge graphs, using a \textit{dynamic triple extraction} method. We supply a flexible number of triples from the knowledge graph as context, tailored to meet the requirements for answering the question. Our model, enriched with knowledge, demonstrates an average improvement of  \textbf{4.75\%} in Exact Match Score over the state-of-the-art on \textbf{three} different KBVQA datasets. Through experiments and analysis, we demonstrate that furnishing variable triples for each question \textit{improves the reasoning capabilities of the language model} in contrast to supplying a fixed number of triples. This is illustrated even for recent large language models. Additionally, we highlight the model's generalization capability by showcasing its SOTA-beating performance on a small dataset, achieved through straightforward fine-tuning.
\end{abstract}


\section{Introduction}\label{sec:intro}
The domain of Knowledge-Based Visual Question Answering (KBVQA) not only utilizes visual information extracted from images, such as object attributes and visual relationships but also integrates supporting facts to facilitate accurate reasoning and answer prediction.
\begin{figure}
    \centering
    \includegraphics[width=0.23 \textwidth, height=0.13\textheight]{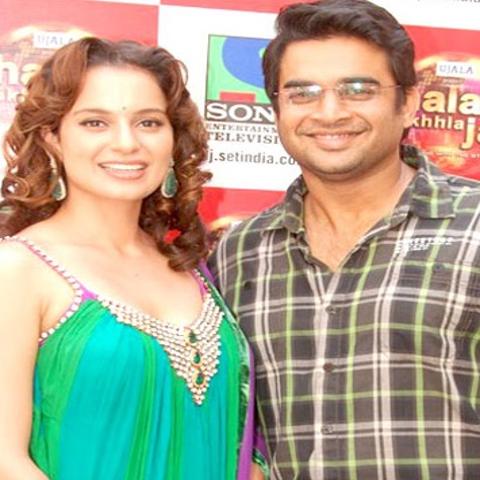}
    \caption{ Example question answerable solely from an image~\citep{Shah_Mishra_Yadati_Talukdar_2019}, without requiring external information.\\
    Question: Who is to the right of R.Madhavan?\\
    Named Entities: [Kangana Ranaut, R. Madhavan]
   }
    \label{fig:my_image2}
\end{figure}\\
\textbf{Motivation: }Recently, large language models (LLMs) like GPT-4~\citep{openai2023gpt4} have garnered attention for their human-like understanding of both images and language, enabling them to tackle KBVQA questions very effectively. However, these LLMs come with a significant drawback: their immense size (around a trillion parameters) poses challenges for offline usage. Additionally, they struggle with user-centric data, such as questions related to named entities within an image. For instance, consider questions like \textit{Who is the person in the middle of the image?} or \textit{What is the age of the person shown in the image?} In such cases, a model should provide specific answers, such as the person’s name or age, rather than generic responses like \textit{man} or \textit{I can’t guess the age}. This could also limit the performance of many IoT applications where real-time user-centric data plays a crucial role.\\
To solve this problem previous works in KBVQA \citep{10.1145/3394171.3413943, garciaolano2021improving, vickers-etal-2021-factuality} used a fixed number of triples from knowledge graphs as additional sources of information to answer the question. Nevertheless, using a fixed number of triples for all questions may lead to either inadequate information or unnecessary noise, potentially resulting in inaccurate predictions. E.g. in Figure~\ref{fig:my_image2}, \textit{Who is to the right of R.Madhavan?} These questions can be answered from image features alone and when additional knowledge is given it introduces noise which often confuses the model and subsequently leads to incorrect predictions. Similarly, some questions require more triples to reach the correct answer, but providing a fixed number of triples can limit the model’s reasoning capabilities due to insufficient information. \\
\textbf{Our Approach: }To address this, we propose a \textbf{dynamic triple filtering} module capable of retrieving a variable number of triples from knowledge graphs as context to answer the questions. We use an established vision language transformer encoder-decoder (OFA) ~\citep{wang2022ofa} model which takes an image, question, and filtered triples as input to predict the desired answer.

\textbf{Our contributions are,}
\begin{enumerate}[nosep]
    \item An approach to Knowledge Based VQA, providing a \textbf{dynamic triple filtering} method that gives question-specific triples instead of a fixed number of ones, serving as context to answer the posed question. The approach surpasses the state-of-the-art (SOTA) on three different KBVQA datasets by at least \textbf{4.12\%}(Section~\ref{sec:expres}, Section~\ref{sec:fvqa} \& Section~\ref{dynamictriples}). 
    
    \item A benchmark across all three datasets. Through a comprehensive evaluation of the VQA model under diverse settings, encompassing both its strengths and weaknesses, we ascertain that the enhanced performance can be attributed to the integration of external knowledge from ConceptNet~\citep{speer2018conceptnet} and WikiData~\citep{10.1145/2629489} in the form of an additional "knowledge vector". (Section~\ref{sec:expres})
    \item An enhanced knowledge base for the CRIC-VQA dataset. This enhancement raises the number of triples from 3,439, as documented in CRIC-VQA~\citep{9905976}, to 99,586 triples 
    (Section~\ref{cricresults})
\end{enumerate}

\section{Related Work}\label{sec:relwork}



\textbf{Knowledge-based VQA:} KBVQA is a recent advancement that incorporates external knowledge along with images and questions to arrive at an answer. There are various datasets published for this purpose. These datasets are mainly of two types-

\textbf{Open Domain Knowledge-Based VQA} involves answering questions that require broad-world knowledge, going beyond what’s directly visible in an image. Several datasets, such as OK-VQA\citep{marino2019okvqa}, A-OKVQA\citep{schwenk2022aokvqa}, and ScienceQA\citep{lu2022learn}, fall into this category. Researchers have tackled this challenge by leveraging various sources of information. Recent work utilizes large language models (LLMs) like GPT-3.5~\citep{gui2022kat,lin2022revive} to retrieve relevant knowledge. Works such as~\citep{khademi-etal-2023-mm,lin2022revive,gui2022kat} found that increasing the diversity of knowledge sources leads to improved accuracy in answering these types of questions.

\textbf{Closed Domain Knowledge-Based VQA} pertains to questions that rely on information from a fixed knowledge base. Datasets like FVQA~\citep{wang2017fvqa}, KVQA~\citep{Shah_Mishra_Yadati_Talukdar_2019}, ViQuae~\citep{10.1145/3477495.3531753}, and CRIC-VQA~\citep{9905976} fall into this category. Some approaches like ~\citep{shevchenko2021reasoning,10.1145/3394171.3413943}, utilize knowledge graphs to retrieve relevant information needed to answer specific questions. Others~\citep{lerner:hal-04384431} have employed a fixed multimodal knowledge base, which combines information from different modalities to provide accurate answers.\\
As user-centric or factual questions require a limited knowledge base to answer a question our work focuses on Closed Knowledge Based VQA. 
In previous works, the MEMNET architecture~\citep{tai2017memnet} was utilized. It retrieved relevant facts from knowledge graphs and then passed them to a BI-LSTM~\citep{huang2015bidirectional} to find the answer. Recent models have leveraged the Vision+Language BERT model~\citep{su2020vlbert} to obtain desired answers. Another approach, proposed by ~\citep{chen2020uniter}, utilizes a BERT-based encoder UNITIER~\citep{devlin-etal-2019-bert} which frames VQA as a classification problem. However, this method has limitations in its applicability to other datasets due to fixed class labels. The latest work, POP-VQA by ~\citep{Sahu_2024_WACV}, employs MT-CNN to retrieve a fixed number of highly relevant facts. These relevant facts, along with questions and images, are then fed into a transformer encoder-decoder model to obtain the desired answer.

\begin{figure*}[t]
  \centering
  \includegraphics[width=0.9\textwidth, height=0.48\textheight]{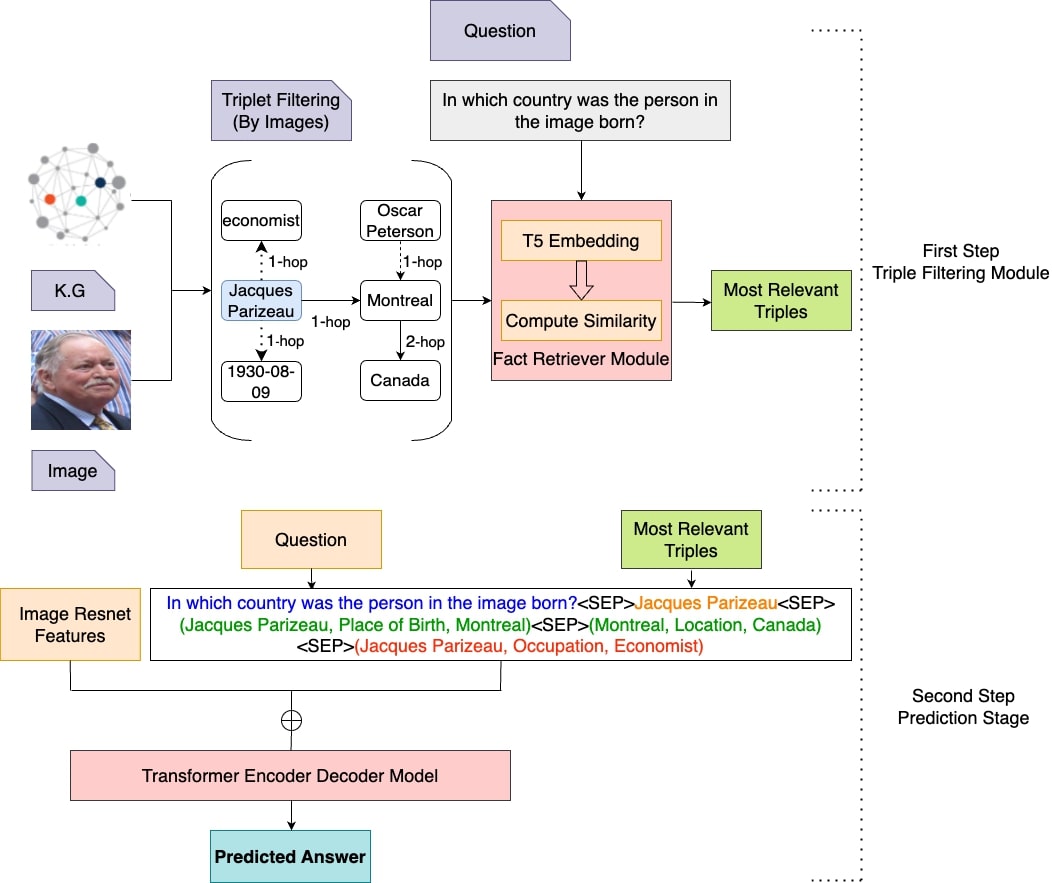}
  \caption{The proposed framework is illustrated in the flow diagram. In the first stage of prediction, triples are filtered based on images, followed by an additional round of filtering based on questions. Finally, the extracted triples in \textcolor{teal}{green} represent useful triples and the triples in \textcolor{red}{red} represent noisy ones. In the second stage of prediction, \textbf{Relevant Triples}, \textbf{Image Resnet Features} and \textbf{Questions}, are fed into a transformer encoder-decoder model (OFA) to generate the predicted answer.$\oplus$ represents the concatenation of all the features to pass it to the transformer encoder-decoder to get the predicted answer.
Irrelevant triples are depicted with dashed lines, while relevant triples, filtered based on images, are represented with bold lines.}
  \label{fig:my_image3}
\end{figure*}

\paragraph{Datasets:}
Three primary datasets, KVQA~\citep{Shah_Mishra_Yadati_Talukdar_2019}, CRIC-VQA~\citep{9905976}, and FVQA~\citep{wang2017fvqa}, are used for Closed Domain Knowledge-Based VQA. KVQA contains 183,000 Q\&A pairs, emphasizing named entity understanding with 18,000 entities across 24,000 images. Conversely, FVQA and CRIC-VQA prioritize commonsense over named entities, with FVQA having 5826 questions and 2190 images, and CRIC-VQA comprising 494K questions and 94K images. We primarily used external knowledge sourced from the Wikidata~\citep{10.1145/2629489} and ConceptNet~\citep{speer2018conceptnet} knowledge graphs to address questions within these datasets.

\section{Our Approach}
Our approach follows a two-stage process to determine the answer for a given question. Let $\mathcal{A}$ be the set of potential answers, $\mathcal{I}$ be the set of images, $i$ be the input image, $\mathcal{Q}$ be the set of questions, and $q$ be the input question. $a^*$ represents the predicted answer where $a^* \in \mathcal{A}$, and $\theta$ represents the learnable parameters of the model.
Then the predicted answer\\
\begin{equation}
a^* = \argmax_{a \in \mathcal{A}} P(a|q, i; \theta)
\end{equation}

Where $P(a|q, i; \theta)$ represents the probability of an answer given a question and the image. $P(a|q, i; \theta)$ is computed in two stages, namely, the triple filtering stage and the prediction stage.

\textbf{Triple Filtering Stage:} Given a question $q$ and an image $i$, we retrieve a set of triples $t^* \subset \mathcal{T}$ using an iterative retrieval mechanism, where $\mathcal{T}$ is the whole set of triples in the knowledge graph.

\begin{equation}
t^* = \bigcup_{t \in \mathcal{T}} (t|(P(t|q, i; \theta_r) >= \lambda))
\end{equation}

here, $\theta_r$ is the set of learned parameters of the fact retriever module, and $\lambda$ is a threshold hyper-parameter.

We integrate multi-hop triples, where we specifically focus on utilizing 2-hop triples for contextual information.

\textbf{Prediction Stage:}
Then we compute the probability of an answer given question, image and relevant triples as:
\begin{equation}
P(a|i,q) = P(a|t^*, i, q; \theta_p)
\end{equation}

here, $\theta_p$ are the learned parameters of the predictor module and $\theta = \theta_p \cup \theta_r$

\subsection{Triple Filtering Module}\label{sec:metho1}
In this module, we extract relevant information from a large-scale knowledge graph to address questions in KBVQA datasets. It involves two distinct steps: 
\subsubsection{Triples Relevant to Entities in Image}\label{expr:alltriples}
Our initial step involves extracting image-relevant triples from a vast knowledge graph, effectively reducing dataset size by eliminating unnecessary information. In KVQA and CRIC-VQA, labels representing named entities or object names in the images are available within the dataset, enabling the extraction of relevant triples by identifying all triples with head or tail entities corresponding to these labels. However, in datasets like FVQA lacking inherent labels, we utilize an alternative method detailed in Section~\ref{sec:fvqa} to extract relevant triples.

\subsubsection{Triples Relevant to Entities in Question}\label{dynamictriples}
From the refined subset of triples obtained from the first step, the module further refines the triple selection by filtering on the question. 

In our approach, inspired by prior work ~\citep{wang-etal-2014-knowledge,9170315,soton480728}, we leverage embedding similarities to find relevant triples. Preceding triple embedding computation, we substitute all named entities with a <MASK> token. This substitution ensures the model prioritizes predicates over named entities, mitigating the extraction of irrelevant triples. For example, in a query like 'Who is to the right of R.Madhavan?', employing <MASK> prevents irrelevant triples like (R.Madhavan, spouse, Sarita Birje) from being extracted to answer the question.
 
\begin{table}[h]
\centering
\begin{adjustbox}{max width=\columnwidth}
\begin{tabular}{|c|c|c|c|c|c|}
\hline
Number of triples & 1 & 3 & 5 & 7 & 9 \\
\hline
Accuracy & 68.95\% & 73.42\% & \textbf{82.7\%} & 82.6\% & 80.20\% \\
\hline
\end{tabular}
\end{adjustbox}
\caption{Exact-match scores when fixed numbers of triples are provided as context for the KVQA dataset.}
\label{tab:varytriples}
\end{table}
In contrast to prior studies, which provided a fixed number of similar triples for answer prediction, our approach introduces a dynamic triplet filtering method. We offer the model a variable number of triples based on a similarity threshold criterion. We include triples with similarity scores equal to or greater than the specified threshold. After observation, we found that a threshold of 0.8 effectively captures nearly all relevant triples needed to answer the given questions. 

The outcomes of the above approaches are presented in Section~\ref{sec:expres}.

\subsection{Prediction Module}
To predict the answer based on an image, question, and triples extracted from the triple filtering module, we employ a transformer encoder-decoder model known as OFA~\citep{wang2022ofa}. The complete architecture is depicted in Figure~\ref{fig:my_image3}, offering a comprehensive overview of our approach.
Due to space constraints, we have given the details of the OFA model in Appendix~\ref{ofamodel}.

\begin{table*}[h]
    \centering
    \tiny
    \renewcommand{\arraystretch}{1.15}
    \resizebox{1.8\columnwidth}{!}{%
    \begin{tabular}{lccccc}
    \toprule
    &&&&\multicolumn{2}{c}{\textbf{OFA(Ours)}}\\
    \textbf{Types of Questions} & \textbf{MEMNET} & \textbf{UNITIER} & \textbf{POP-VQA} & \textbf{Single-Hop} & \textbf{Multi-Hop} \\
    \midrule
    1-Hop & 61.00\% & 65.70\% & \textbf{89.80\%} & 84.25\% & \underline{86.04\%} \\ 
    Boolean & 75.10\% & 94.60\% & 95.70\% & \underline{96.89\%} & \textbf{97.17\%}\\
    Comparison & 50.50\% & \underline{90.40\%} & 89.60\% & \textbf{90.82\%} & 90.15\% \\
    Counting & 49.50\% & 79.40\% & 73.20\% & \underline{90.08\%} & \textbf{90.32\%} \\
    Intersection & 72.50\% & 79.40\% & 72.30\% & \underline{87.07\%} & \textbf{89.03\%}\\
    Multi-Entity & 43.50\% & 77.10\% & \textbf{94.90\%} & 84.01\% & \underline{88.53\%} \\
    Multi-Relation & 45.20\% & 75.20\% & \textbf{93.27\%} & 90.10\% & \underline{90.77\%} \\
    Spatial & 48.10\% & 21.20\% & 83.89\% & \underline{92.70\%} & \textbf{94.50\%} \\
    Subtraction & \textbf{40.50\%} & 34.40\% & 37.00\% & 32.50\% & \underline{40.20\%} \\
    \midrule
    \textbf{Average Scores} & 53.98\% & 68.60\% & 81.07\% & \underline{83.15\%} & \textbf{85.19\%} \\ \bottomrule
    \end{tabular}%
    }
    \caption{\textbf{Results on KVQA}~\citep{Shah_Mishra_Yadati_Talukdar_2019}. Exact match scores for various question types. These scores are obtained in a setting where triples are filtered based on both the questions and the images, and the number of triples varies according to a similarity threshold. We show a comparison of our results with the performance of previous baseline models,
MEMNET~\citep{tai2017memnet}, UNITIER~\citep{chen2020uniter} and POP-VQA~\citep{Sahu_2024_WACV}, on the KVQA test set. Bold and \underline{underline} indicate the best and second-best scores. Overall our model outperforms the baseline across the test set and most of the classes.}
    \label{tab:mytable}
\end{table*}%

Algorithm~\ref{exp:algo} outlines the high-level process of retrieving relevant triples and making answer predictions as shown in Appendix~\ref{algo1}.

\section{Experimental Setup \& Results}\label{sec:expres}
In this section, we explain the results of KVQA and CRIC-VQA datasets. 
\subsection{Results on KVQA dataset}\label{kvqaresults}

Table~\ref{tab:mytable} displays the results on the KVQA dataset. Overall, our model exhibits superior performance compared to baseline models and surpasses them in the majority of categories. The KVQA dataset includes 12 classes. However, prior research only made comparisons across 9 classes. Therefore, we also present our results for these 9 classes for a fair comparison. Our model achieves an average score of ~\textbf{85.19\%} on the KVQA dataset which is ~\textbf{4.12\%} better than the SOTA model POP-VQA. We have also included the results for all 12 classes in the Appendix~\ref{sec:appendixaddkvqa}. However, we do encounter a major shortfall in the multi-entity class, where our performance is ~\textbf{6\%} lower than the current SOTA model, POP-VQA. We attribute this performance gap to the fact that the POP-VQA model is specifically trained for single-hop question-answering contexts. At the same time, our approach incorporates multi-hop triples, potentially introducing additional noise that could affect prediction accuracy.

 Previous approaches mainly rely on a fixed number of the most similar triples as context for predictions, our approach employs dynamic filtering, enhancing the model's reasoning capabilities, as shown in Table ~\ref{tab:my-table1}.
\subsubsection{Ablation Results}
We demonstrate the efficacy of our dynamic filtering method across various settings. Typically, inputs consist of Image Features, Questions, Named Entities, and Context, separated by <SEP> tokens and fed into the transformer encoder. Answers are generated by the transformer decoder. The context is structured as a sequence of triples, labelled as $triple_1$ <SEP> $triple_2$ <SEP> $triple_3$... <SEP> $triple_n$, with triples are in the form (head, relation, tail). These settings include:
\begin{enumerate}[nosep]
    \item \textbf{No External Knowledge} (Table~\ref{tab:my-table1}, Row 1): In this setting, we provided image features and questions without any context.
    \item \textbf{Triples Related to Images} (Table~\ref{tab:my-table1}, Row 2): Here, we included all triples associated with named entities in the image.
    \item \textbf{Triple Filtering Based on Questions:}\\
    In this context, there exist two configurations,\\
    \textbf{Fixed Number of Triples} (Table~\ref{tab:my-table1}, Row 4\&6): We choose a fixed number of $\text{top-5}$ triples with the highest similarity scores. While we experimented by varying numbers of triples, as depicted in Table~\ref{tab:varytriples}, we observed that providing top-5 triples as context yielded the highest accuracy.\\
    \textbf{Dynamic Number of Triples with Similarity Threshold} (Table~\ref{tab:my-table1}, Row 3\&5): We selected all triples with a similarity greater than or equal to 0.8.
\end{enumerate}
For comparison with baselines on the KVQA dataset, we conducted evaluations on the OFA large model, utilizing a dynamic number of multi-hop triples to determine accuracy across various question classes. 
\begin{table}[h]
    \centering
    \resizebox{0.95\columnwidth}{!}{%
    \begin{tabular}{lcc}
    \toprule
    \textbf{Models}&\textbf{Base}&\textbf{Large}\\
    \midrule
    OFA+Image & 62.70\% & 76.70\%\\
    OFA+Image+All Triples & 72.00\% & 73.67\%\\
\begin{tabular}[c]{@{}l@{}}OFA+Image+Filtered Triples(Dynamic)\\        (Single-Hop)\end{tabular} & 83.65\% & 85.35\% \\
\begin{tabular}[c]{@{}l@{}}OFA+Image+Filtered Triples(Top-5)\\          (Single-Hop)\end{tabular} & 82.45\% & 83.20\%  \\
\begin{tabular}[c]{@{}l@{}}OFA+Image+Filtered Triples(Dynamic)\\         (Multi-Hop)\end{tabular} & \textbf{85.15\%} & \textbf{87.55\%} \\
\begin{tabular}[c]{@{}l@{}}OFA+Image+Filtered Triples(Top-5)\\          (Multi-Hop)\end{tabular} & 83.57\% & 82.70\%\\
    \bottomrule
    \end{tabular}%
    }
    \caption{\textbf{Ablation Results on the KVQA Dataset}. All Triples (Row 2) refers to image-only triple filtering, Filtered Triples involve filtering based on both question and image. In the second approach, two settings are considered: 1) Fixed triples with Top-5 context and 2) Dynamic triples with a similarity threshold. Bold indicates best scores.}
    \label{tab:my-table1}
\end{table}%
\begin{table}[t]
    \centering
    \tiny
    \resizebox{0.7\columnwidth}{!}{%
    \begin{tabular}{lc}
    \toprule
    \textbf{Models} & \textbf{Accuracy}\\
    \midrule
    Q-Only GRU & 55.18\% \\
    Q-Only-BERT & 59.03\% \\
    SAN & 63.98\% \\
    Bottom-Up+latt & 62.39\% \\
    MAC-CS & 69.65\% \\
    NMN-CS & 68.96\% \\
    Memory-VQA+latt & 66.93\% \\
    VILBERT+latt & 77.54\% \\
    VILBERT+ERNIE+latt & 79.85\% \\
    \midrule
    \multicolumn{2}{c}{\textbf{Ours}}\\
    \midrule
    OFA Base (Fixed) & 76.17\% \\
    OFA Large (Fixed) & 79.28\% \\
    OFA Base (Dynamic) & 81.85\% \\
    OFA Large (Dynamic) & \textbf{85.80\%} \\
    \bottomrule
    \end{tabular}%
    }
    \caption{\textbf{Results on CRIC-VQA}~\citep{gao2021cric}. Exact match scores for various baselines as well as our model. Fixed denotes fixed number of triples with Top-5 context, and dynamic denotes variable triples with a similarity threshold.}
    \label{tab:cricvqatable}
\end{table}
\subsection{Results on CRIC-VQA dataset}\label{cricresults}
Table~\ref{tab:cricvqatable} presents the results obtained on the CRIC-VQA dataset, which features factual questions requiring commonsense reasoning. Due to the dataset's is not open source, there has been limited prior research. In our study, we compared our method with nine baseline models outlined in ~\citep{9905976}. Our approach achieves an accuracy of \textbf{85.80\%}, surpassing the SOTA model VILBERT+ERNIE+latt by \textbf{5.95\%}. As the number of baselines is high, we explain each in the Appendix~\ref{appendix:cricbaselines}.

The CRIC-VQA dataset possesses a relatively small knowledge base, containing approximately ~\textbf{3,400} triples. To demonstrate the effectiveness of our method, we expanded the knowledge base using ConceptNet. Our augmentation involved incorporating all triples related to the objects depicted in the images, ensuring alignment with either the head or tail entity corresponding to the object label. Consequently, we augmented the knowledge base to a substantial \textbf{99,586} triples. This significant increase presents challenges in extracting relevant knowledge for question-answering tasks. Given the dataset's size and computational constraints, our experiments primarily focused on filtering context based on both images and questions.

Due to space constraints, training details are present in Appendix~\ref{appendix:traindetails}.
\section{Generalisation Capability}\label{sec:fvqa}
\begin{figure}
    \centering
    \includegraphics[width=0.42 \textwidth, height=0.22\textheight]{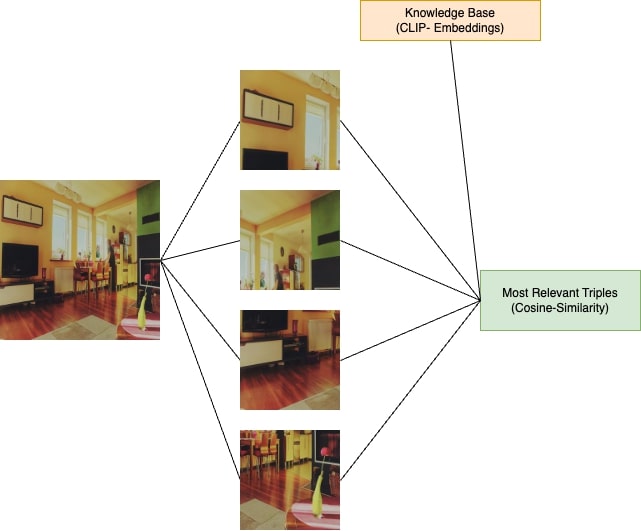}
    \caption{Splitting the image into four patches to extract relevant triples.}
    \label{fig:splitimage}
\end{figure}

We demonstrate our model's generalization capability by fine-tuning it on the FVQA dataset following pretraining on the KVQA dataset. The primary challenge with the FVQA dataset is the absence of object labels within the dataset itself. So extracting image-relevant triples directly from the knowledge graph by matching the head or tail entity is not possible. We fine-tuned the CLIP model~\citep{radford2021learning} to get image-relevant triples. We have included details of fine-tuning the CLIP model in Appendix~\ref{CLIPfine-tuning}.
To find image-relevant triples we calculate the CLIP embedding~\citep{radford2021learning} for each triple. To ensure we extract relevant triples for small objects in the image, we divide the image into four equal-sized patches and compute the CLIP embedding for each patch. When examining the entire image without dividing it into patches, important details related to small objects (such as the flower vase) as shown in Figure~\ref{fig:splitimage} might be overlooked. Cosine similarity between patch embeddings and all triples is calculated, selecting those with a similarity above 0.8 for each patch. 
For extracting the triples relevant to the question we use the same approach as explained in Section~\ref{dynamictriples}.
\begin{table}[t]
    \centering
    \resizebox{\columnwidth}{!}{%
    \begin{tabular}{lc}
    \toprule
    \textbf{Models} & \textbf{Accuracy}\\
    \midrule 
   Human & 77.99\% \\
   FVQA~\citep{10.1109/TPAMI.2017.2754246} & 56.91\% \\
   ZS-FVQA~\citep{chen2021zeroshot} & 58.27\% \\
   FVQA (Ensemble)~\citep{10.1109/TPAMI.2017.2754246} & 58.76\% \\
    MM-Reasoner (Ensemble)~\citep{khademi-etal-2023-mm}  & 61.10\% \\
    \midrule
    \multicolumn{2}{c}{\textbf{Ours}}\\
    \midrule
    OFA Base(Ours) & 54.00\% \\
    OFA Large(Ours) & \textbf{65.28\%} \\
    \bottomrule
    \end{tabular}%
    }
    \caption{\textbf{Results on FVQA}. Exact match scores for various baselines as well as our model. Utilized pre-trained model on KVQA dataset under dynamic multi-hop setting (Table~\ref{tab:my-table1}). The inference is done while providing the dynamic number of triples as context.}
    \label{tab:fvqatable}
\end{table}
Our approach achieves an accuracy of ~\textbf{65.28\%}, surpassing the SOTA by ~\textbf{4.28\%}. In Table~\ref{tab:fvqatable}, we compare our work with previous baselines, particularly those that do not consider named entities when extracting relevant triples from the knowledge base. This improvement is attributed to our model's ability to eliminate irrelevant noise introduced by external context, unlike the SOTA model MM-Reasoner, which integrates context from diverse sources such as image captions, GPT4 and many more.\\
\textbf{Ablation Results :}
We explore various settings to demonstrate the effectiveness of incorporating a dynamic number of triples during pretraining or fine-tuning, as depicted in Table~\ref{tab:mytable5}. The table showcases results with and without fine-tuning the FVQA dataset, presenting different contextual information during these processes. Notably, we find that utilizing a dynamic number of triples leads to a \textbf{12\%} performance improvement compared to a fixed number. 
We also include results without image segmentation into patches and results obtained by identifying triples relevant to objects in the image. For this, we utilize the object detection model~\citep{wu2019detectron2} to obtain bounding boxes for objects and extract the most relevant triples for each box.
 For a comprehensive understanding of these approaches, including detailed results, please refer to Appendix~\ref{additresFVQA}. Despite the distinct domains between the fine-tuning dataset (FVQA) and the pretraining dataset (KVQA), our model exhibits strong performance and generalizability across different domains.
\begin{table}[h]
    \centering
    \resizebox{1\columnwidth}{!}{%
    \begin{tabular}{lccccc}
    \toprule
    &\multicolumn{2}{c}{\textbf{Context-Type}}&&\\
    \textbf{Models} & \textbf{Pre-training} & \textbf{Inference} & \textbf{$\sim$ FT} & \textbf{FT}\\
    \midrule
    Base & fixed & fixed & 20.35\% & 43.00\% \\
    Base & fixed & dynamic & 21.90\% & 30.00\% \\
    Large & fixed & fixed & 36.48\% & 39.00\% \\
    Large & fixed & dynamic & 38.70\% & 50.00\% \\
    Base & dynamic & fixed & 34.50\% & 41.51\% \\
    Base & dynamic & dynamic & 40.14\% & 54.00\% \\
    Large & dynamic & fixed & 43.50\% & 58.00\% \\
    Large & dynamic & dynamic & \textbf{47.00\%} & \textbf{65.28\%} \\
    \bottomrule
    \end{tabular}%
    }
    \caption{\textbf{Ablation results for FVQA dataset}: Exact match scores comparing fine-tuned (FT) and non-fine-tuned ($\sim$FT) models, pre-trained on the KVQA dataset. The pre-training context type specifies how the model was trained on the KVQA dataset, while the inference context type indicates the settings for fine-tuning and inference on the FVQA dataset.}
    \label{tab:mytable5}
\end{table}
\section{Relevance of Knowledge in the Context of MLLMs}\label{prompting}
Given the extensive training of MLLMs on vast datasets, it's natural to assume that external knowledge might not be essential for using them in tasks like KB-VQA. Recently, there have been extensive discussions about whether Multimodal Large Language Models (MLLMs), trained on large datasets, can answer KB-VQA questions based solely on their internal knowledge or if external information is necessary. In this section, we illustrate that relying solely on the implicit knowledge within MLLMs is insufficient for addressing such questions. Additionally, we'll demonstrate the effectiveness of our knowledge retrieval method by evaluating its performance with a contemporary vision language model.
\subsection{Zero-shot Evaluation on the LLAVA model}
 We conducted experiments with the MLLM llava-v1.6-vicuna-13b\footnote{https://huggingface.co/liuhaotian/llava-v1.6-vicuna-13b}, prompting it to generate responses to zero-shot image prompts under two conditions: one without external knowledge and the other with external knowledge obtained through our dynamic triple retrieval module. The prompts for both conditions are detailed in Appendix~\ref{appendix:llavaprompt}.
\begin{table}[h]
    \centering
    \resizebox{0.95\columnwidth}{!}{%
    \begin{tabular}{lcc}
    \toprule
    \textbf{Dataset}&\textbf{Without External Knowledge}&\textbf{With External Knowledge}\\
    \midrule
    KVQA & 55.20\% & 64.50\%\\
    CRIC-VQA & 58.60\% & 69.40\%\\
    \bottomrule
    \end{tabular}%
    }
    \caption{\textbf{Zero-shot results on LLAVA model: }Exact Match scores achieved by the llava-v1.6-vicuna-13b model. The results are reported for two settings: (1) without providing any prior knowledge and (2) with the inclusion of knowledge in the form of triples, alongside questions and named entities found in the images.}
    \label{tab:llavawwoknow}
\end{table}%
For evaluation, we computed exact match scores by comparing generated answers with correct ones. 
To determine whether discrepancies arose from the model's output or it is exact match metric issues, we performed a qualitative analysis on 200 incorrect samples. Additional information regarding this is provided in Appendix~\ref{humaneval}.\\
The results of the above approach are shown in Table~\ref{tab:llavawwoknow}. We can observe incorporating explicit external knowledge increased accuracy by approximately \textbf{10.05\%}. The significant increase in accuracy demonstrates how even large language models benefit from incorporating external knowledge.
\subsection{Finetuning LLAVA on KVQA dataset}
Here we demonstrate the effectiveness of our approach while finetuning LLAVA for the KB-VQA task. 
We fine-tuned LLAVA in three different settings:
\begin{enumerate}[nosep]
\item \textbf{Without Knowledge}: Here no context is provided to answer the question.
\item \textbf{With Fixed Knowledge}: Here top-5 most similar triples are provided as context to answer the question.
\item \textbf{With Dynamic Knowledge}: Here triples obtained from the dynamic triple retrieval module are provided to answer the question.
\end{enumerate}
The results are shown in Table~\ref{tab:my-tablellavatech}. There is a substantial increase in accuracy \textbf{15.7\%}, on the KVQA test set when contextual information is integrated while fine tuning. This accuracy gap further widens to \textbf{20.2\%} when the triples are filtered using our dynamic filtering approach. These results underscore the effectiveness of our approach, even with recent models, significantly enhancing prediction accuracy.
\begin{table}[h]
    \centering
    \resizebox{0.95\columnwidth}{!}{%
    \begin{tabular}{lc}
    \toprule
    \textbf{Technique}&\textbf{Accuracy}\\
    \midrule
    LLAVA+Labels & 72.40\%\\
    LLAVA+Labels+Knowledge (fixed) & 88.10\%\\
    LLAVA+Labels+Knowledge (dynamic) & \textbf{92.60\%}\\
    \bottomrule
    \end{tabular}%
    }
    \caption{\textbf{Results of LLAVA on KVQA Dataset:} Exact Match scores achieved by the LLAVA model after fine-tuning on the KVQA Dataset for the KB-VQA task. In this context, \textbf{labels} refer to named entities detected within the images, while \textbf{knowledge} indicates external information supplied either statically or dynamically.}
    \label{tab:my-tablellavatech}
\end{table}
\section{Qualitative Analysis}\label{sec:quali}
We'll illustrate how integrating knowledge boosts the OFA model's predictive power while efficiently filtering noise enhances its reasoning capability. We choose some samples as shown in Table~\ref{my-labeltab4}. The first row shows that giving extra information helps the model make correct predictions for simple questions with straightforward answers. In this case, the correct answer is obtained whether or not we filter the knowledge triples.
\begin{table}[t]
\small
\centering
\setlength{\extrarowheight}{0.5em}
\begin{tabular}{>{\tiny}m{1.0cm}|>{\tiny}m{1.4cm}|>{\tiny}m{0.5cm}|>{\tiny}m{0.5cm}|>{\tiny}m{0.5cm}|>{\tiny}m{0.5cm}}
\textbf{Image} & \textbf{Question} & \textbf{Truth Value} & \textbf{No Triples} & \textbf{All Triples} & \textbf{Filtered Triples} \\ \hline
\includegraphics[scale=0.09,valign=c]{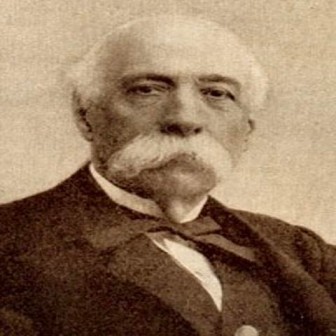} & Is the person in the image a politician? & No & Yes & No & No \\[1em] \hline
\includegraphics[scale=0.09,valign=c]{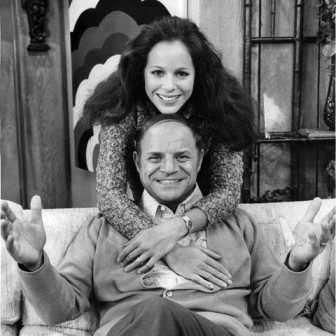} & For how many years did the person in the image live? & 83 & 72 & 82 & 83 \\[1em] \hline
\includegraphics[scale=0.09,valign=c]{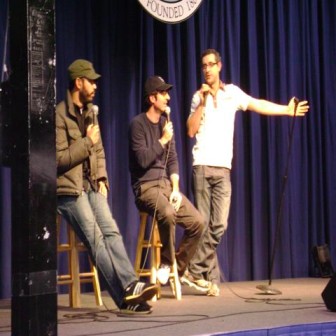} & Were all the people in the image born in the same country? & No & Yes & Yes & No \\[1em] \hline
\includegraphics[scale=0.09,valign=c]{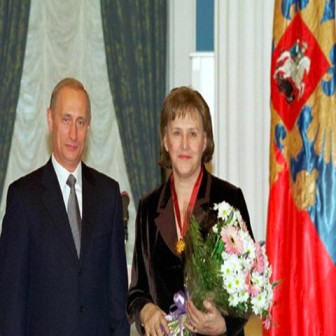} & Who among the people in the image ever married Vladimir Soshalsky? & Person on the left & Person on the right & Person on the right & Person on the right \\[1em] \hline
\end{tabular}
\caption{Qualitative analysis, which presents instances from
the dataset and their answer predictions with and without the presence of triples in the input.}
\label{my-labeltab4}
\end{table}
Rows 2 and 3 demonstrate that providing all triples filtered based on image and not question introduces irrelevant knowledge (noise), resulting in inaccurate predictions. However, filtering triples based on questions leads to correct answers. These complex questions include a single-hop subtraction (Row 2) and a multi-hop boolean query (Row 3), requiring noise removal for accurate predictions.

In the final example, regardless of whether knowledge triples are supplied or not, the model produces incorrect answers. This question falls under the spatial and multi-hop categories, requiring the model to make inferences based on both image features and external knowledge. One possible explanation for the inaccurate predictions could be the model's insufficient training to a broad range of questions within these complex categories, hindering its ability to reason effectively in such scenarios. Due to space constraints, we provide these and some more examples in Table~\ref{my-labellasttable}.
\section{Conclusion}\label{sec:concl}
We presented a novel approach for KBVQA, utilizing a dynamic triple-filtering module to extract external context from knowledge graphs. Our method outperforms the SOTA on three different KBVQA datasets, achieving an average improvement of \textbf{4.75\%}. We demonstrated that providing the model with a varying number of triples during pre-training or fine-tuning enhances its reasoning capabilities compared to a fixed number of triples. We also showcased the generalization capability of our approach by achieving SOTA performance on a small dataset using a model trained on a completely different domain. Furthermore, we demonstrated that large MLLMs also require external knowledge for accurate responses. Finally, we also demonstrated the effectiveness of our approach on the latest MLLM LLAVA, emphasizing that providing a dynamic number of triples improves accuracy by \textbf{20\%} as compared to static ones. The key insight is that dynamically determining the number of relevant triples in the context eliminates noise, resulting in more precise predictions. We also discuss some future explorations in Appendix~\ref{appendix:future}.

\newpage
\textbf{Limitations:}
Our approach requires calculating the similarity between the question and all the triples associated with the images to identify the most relevant ones. This process can become computationally intensive, especially when dealing with a substantial number of triples, resulting in longer prediction times.

\bibliography{custom}

\begin{thebibliography}{36}
\expandafter\ifx\csname natexlab\endcsname\relax\def\natexlab#1{#1}\fi

\bibitem[{Bostrom and Durrett(2020)}]{bostrom2020byte}
Kaj Bostrom and Greg Durrett. 2020.
\newblock \href {http://arxiv.org/abs/2004.03720} {Byte pair encoding is suboptimal for language model pretraining}.

\bibitem[{Chen et~al.(2020)Chen, Li, Yu, Kholy, Ahmed, Gan, Cheng, and Liu}]{chen2020uniter}
Yen-Chun Chen, Linjie Li, Licheng Yu, Ahmed~El Kholy, Faisal Ahmed, Zhe Gan, Yu~Cheng, and Jingjing Liu. 2020.
\newblock \href {http://arxiv.org/abs/1909.11740} {Uniter: Universal image-text representation learning}.

\bibitem[{Chen et~al.(2021)Chen, Chen, Geng, Pan, Yuan, and Chen}]{chen2021zeroshot}
Zhuo Chen, Jiaoyan Chen, Yuxia Geng, Jeff~Z. Pan, Zonggang Yuan, and Huajun Chen. 2021.
\newblock \href {http://arxiv.org/abs/2107.05348} {Zero-shot visual question answering using knowledge graph}.

\bibitem[{Devlin et~al.(2019)Devlin, Chang, Lee, and Toutanova}]{devlin-etal-2019-bert}
Jacob Devlin, Ming-Wei Chang, Kenton Lee, and Kristina Toutanova. 2019.
\newblock \href {https://doi.org/10.18653/v1/N19-1423} {{BERT}: Pre-training of deep bidirectional transformers for language understanding}.
\newblock In \emph{Proceedings of the 2019 Conference of the North {A}merican Chapter of the Association for Computational Linguistics: Human Language Technologies, Volume 1 (Long and Short Papers)}, pages 4171--4186, Minneapolis, Minnesota. Association for Computational Linguistics.

\bibitem[{Gao et~al.(2023)Gao, Wang, Shan, and Chen}]{9905976}
D.~Gao, R.~Wang, S.~Shan, and X.~Chen. 2023.
\newblock \href {https://doi.org/10.1109/TPAMI.2022.3210780} {Cric: A vqa dataset for compositional reasoning on vision and commonsense}.
\newblock \emph{IEEE Transactions on Pattern Analysis \&amp; Machine Intelligence}, 45(05):5561--5578.

\bibitem[{Gao et~al.(2021)Gao, Wang, Shan, and Chen}]{gao2021cric}
Difei Gao, Ruiping Wang, Shiguang Shan, and Xilin Chen. 2021.
\newblock \href {http://arxiv.org/abs/1908.02962} {Cric: A vqa dataset for compositional reasoning on vision and commonsense}.

\bibitem[{Garcia-Olano et~al.(2021)Garcia-Olano, Onoe, and Ghosh}]{garciaolano2021improving}
Diego Garcia-Olano, Yasumasa Onoe, and Joydeep Ghosh. 2021.
\newblock \href {http://arxiv.org/abs/2112.06888} {Improving and diagnosing knowledge-based visual question answering via entity enhanced knowledge injection}.

\bibitem[{Gui et~al.(2022)Gui, Wang, Huang, Hauptmann, Bisk, and Gao}]{gui2022kat}
Liangke Gui, Borui Wang, Qiuyuan Huang, Alex Hauptmann, Yonatan Bisk, and Jianfeng Gao. 2022.
\newblock \href {http://arxiv.org/abs/2112.08614} {Kat: A knowledge augmented transformer for vision-and-language}.

\bibitem[{He et~al.(2015)He, Zhang, Ren, and Sun}]{he2015deep}
Kaiming He, Xiangyu Zhang, Shaoqing Ren, and Jian Sun. 2015.
\newblock \href {http://arxiv.org/abs/1512.03385} {Deep residual learning for image recognition}.

\bibitem[{Huang et~al.(2015)Huang, Xu, and Yu}]{huang2015bidirectional}
Zhiheng Huang, Wei Xu, and Kai Yu. 2015.
\newblock \href {http://arxiv.org/abs/1508.01991} {Bidirectional lstm-crf models for sequence tagging}.

\bibitem[{Khademi et~al.(2023)Khademi, Yang, Frujeri, and Zhu}]{khademi-etal-2023-mm}
Mahmoud Khademi, Ziyi Yang, Felipe Frujeri, and Chenguang Zhu. 2023.
\newblock \href {https://doi.org/10.18653/v1/2023.findings-emnlp.437} {{MM}-reasoner: A multi-modal knowledge-aware framework for knowledge-based visual question answering}.
\newblock In \emph{Findings of the Association for Computational Linguistics: EMNLP 2023}, pages 6571--6581, Singapore. Association for Computational Linguistics.

\bibitem[{Lerner et~al.(2024)Lerner, Ferret, and Guinaudeau}]{lerner:hal-04384431}
Paul Lerner, Olivier Ferret, and Camille Guinaudeau. 2024.
\newblock \href {https://hal.science/hal-04384431} {{Cross-modal Retrieval for Knowledge-based Visual Question Answering}}.
\newblock Working paper or preprint.

\bibitem[{Lerner et~al.(2022)Lerner, Ferret, Guinaudeau, Le~Borgne, Besan\c{c}on, Moreno, and Lov\'{o}n~Melgarejo}]{10.1145/3477495.3531753}
Paul Lerner, Olivier Ferret, Camille Guinaudeau, Herv\'{e} Le~Borgne, Romaric Besan\c{c}on, Jose~G. Moreno, and Jes\'{u}s Lov\'{o}n~Melgarejo. 2022.
\newblock \href {https://doi.org/10.1145/3477495.3531753} {Viquae, a dataset for knowledge-based visual question answering about named entities}.
\newblock In \emph{Proceedings of the 45th International ACM SIGIR Conference on Research and Development in Information Retrieval}, SIGIR '22, page 3108–3120, New York, NY, USA. Association for Computing Machinery.

\bibitem[{Li et~al.(2020)Li, Wang, and Zhu}]{10.1145/3394171.3413943}
Guohao Li, Xin Wang, and Wenwu Zhu. 2020.
\newblock \href {https://doi.org/10.1145/3394171.3413943} {Boosting visual question answering with context-aware knowledge aggregation}.
\newblock In \emph{Proceedings of the 28th ACM International Conference on Multimedia}, MM '20, page 1227–1235, New York, NY, USA. Association for Computing Machinery.

\bibitem[{Lin et~al.(2022)Lin, Xie, Chen, Xu, Zhu, and Yuan}]{lin2022revive}
Yuanze Lin, Yujia Xie, Dongdong Chen, Yichong Xu, Chenguang Zhu, and Lu~Yuan. 2022.
\newblock \href {http://arxiv.org/abs/2206.01201} {Revive: Regional visual representation matters in knowledge-based visual question answering}.

\bibitem[{Lu et~al.(2022)Lu, Mishra, Xia, Qiu, Chang, Zhu, Tafjord, Clark, and Kalyan}]{lu2022learn}
Pan Lu, Swaroop Mishra, Tony Xia, Liang Qiu, Kai-Wei Chang, Song-Chun Zhu, Oyvind Tafjord, Peter Clark, and Ashwin Kalyan. 2022.
\newblock Learn to explain: Multimodal reasoning via thought chains for science question answering.
\newblock In \emph{The 36th Conference on Neural Information Processing Systems (NeurIPS)}.

\bibitem[{Ma et~al.(2019)Ma, Teng, Zhong, and MA}]{9170315}
Minbo Ma, Fei Teng, Wen Zhong, and Zheng MA. 2019.
\newblock \href {https://doi.org/10.1109/ISKE47853.2019.9170315} {A sentence-rcnn embedding model for knowledge graph completion}.
\newblock In \emph{2019 IEEE 14th International Conference on Intelligent Systems and Knowledge Engineering (ISKE)}, pages 484--490.

\bibitem[{Marino et~al.(2019)Marino, Rastegari, Farhadi, and Mottaghi}]{marino2019okvqa}
Kenneth Marino, Mohammad Rastegari, Ali Farhadi, and Roozbeh Mottaghi. 2019.
\newblock \href {http://arxiv.org/abs/1906.00067} {Ok-vqa: A visual question answering benchmark requiring external knowledge}.

\bibitem[{Nayyeri et~al.(2023)Nayyeri, Wang, Akter, Alam, Rony, Lehmann, and Staab}]{soton480728}
Mojtaba Nayyeri, Zihao Wang, Mst.~Mahfuja Akter, Mirza~Mohtashim Alam, Md~Rashad Al~Hasan Rony, Jens Lehmann, and Steffen Staab. 2023.
\newblock \href {https://eprints.soton.ac.uk/480728/} {Integrating knowledge graph embeddings and pre-trained language models in hypercomplex spaces}.
\newblock In \emph{22nd International Semantic Web Conference (06/11/23 - 10/11/23)}.

\bibitem[{OpenAI et~al.(2023)OpenAI, :, Achiam, Adler, Agarwal, Ahmad, Akkaya et~al.}]{openai2023gpt4}
OpenAI, :, Josh Achiam, Steven Adler, Sandhini Agarwal, Lama Ahmad, Ilge Akkaya, et~al. 2023.
\newblock \href {http://arxiv.org/abs/2303.08774} {Gpt-4 technical report}.

\bibitem[{Radford et~al.(2021)Radford, Kim, Hallacy, Ramesh, Goh, Agarwal, Sastry, Askell, Mishkin, Clark, Krueger, and Sutskever}]{radford2021learning}
Alec Radford, Jong~Wook Kim, Chris Hallacy, Aditya Ramesh, Gabriel Goh, Sandhini Agarwal, Girish Sastry, Amanda Askell, Pamela Mishkin, Jack Clark, Gretchen Krueger, and Ilya Sutskever. 2021.
\newblock \href {http://arxiv.org/abs/2103.00020} {Learning transferable visual models from natural language supervision}.

\bibitem[{Raffel et~al.(2023)Raffel, Shazeer, Roberts, Lee, Narang, Matena, Zhou, Li, and Liu}]{raffel2023exploring}
Colin Raffel, Noam Shazeer, Adam Roberts, Katherine Lee, Sharan Narang, Michael Matena, Yanqi Zhou, Wei Li, and Peter~J. Liu. 2023.
\newblock \href {http://arxiv.org/abs/1910.10683} {Exploring the limits of transfer learning with a unified text-to-text transformer}.

\bibitem[{Sahu et~al.(2024)Sahu, Raut, Samant, Gorijala, Lakshminarayanan, and Bhaskar}]{Sahu_2024_WACV}
Pragya~Paramita Sahu, Abhishek Raut, Jagdish~Singh Samant, Mahesh Gorijala, Vignesh Lakshminarayanan, and Pinaki Bhaskar. 2024.
\newblock Pop-vqa - privacy preserving, on-device, personalized visual question answering.
\newblock In \emph{Proceedings of the IEEE/CVF Winter Conference on Applications of Computer Vision (WACV)}, pages 8470--8479.

\bibitem[{Schwenk et~al.(2022)Schwenk, Khandelwal, Clark, Marino, and Mottaghi}]{schwenk2022aokvqa}
Dustin Schwenk, Apoorv Khandelwal, Christopher Clark, Kenneth Marino, and Roozbeh Mottaghi. 2022.
\newblock \href {http://arxiv.org/abs/2206.01718} {A-okvqa: A benchmark for visual question answering using world knowledge}.

\bibitem[{Shah et~al.(2019)Shah, Mishra, Yadati, and Talukdar}]{Shah_Mishra_Yadati_Talukdar_2019}
Sanket Shah, Anand Mishra, Naganand Yadati, and Partha~Pratim Talukdar. 2019.
\newblock \href {https://doi.org/10.1609/aaai.v33i01.33018876} {Kvqa: Knowledge-aware visual question answering}.
\newblock \emph{Proceedings of the AAAI Conference on Artificial Intelligence}, 33(01):8876--8884.

\bibitem[{Shevchenko et~al.(2021)Shevchenko, Teney, Dick, and van~den Hengel}]{shevchenko2021reasoning}
Violetta Shevchenko, Damien Teney, Anthony Dick, and Anton van~den Hengel. 2021.
\newblock \href {http://arxiv.org/abs/2101.06013} {Reasoning over vision and language: Exploring the benefits of supplemental knowledge}.

\bibitem[{Speer et~al.(2018)Speer, Chin, and Havasi}]{speer2018conceptnet}
Robyn Speer, Joshua Chin, and Catherine Havasi. 2018.
\newblock \href {http://arxiv.org/abs/1612.03975} {Conceptnet 5.5: An open multilingual graph of general knowledge}.

\bibitem[{Su et~al.(2020)Su, Zhu, Cao, Li, Lu, Wei, and Dai}]{su2020vlbert}
Weijie Su, Xizhou Zhu, Yue Cao, Bin Li, Lewei Lu, Furu Wei, and Jifeng Dai. 2020.
\newblock \href {http://arxiv.org/abs/1908.08530} {Vl-bert: Pre-training of generic visual-linguistic representations}.

\bibitem[{Tai et~al.(2017)Tai, Yang, Liu, and Xu}]{tai2017memnet}
Ying Tai, Jian Yang, Xiaoming Liu, and Chunyan Xu. 2017.
\newblock \href {http://arxiv.org/abs/1708.02209} {Memnet: A persistent memory network for image restoration}.

\bibitem[{Vickers et~al.(2021)Vickers, Aletras, Monti, and Barrault}]{vickers-etal-2021-factuality}
Peter Vickers, Nikolaos Aletras, Emilio Monti, and Lo{\"\i}c Barrault. 2021.
\newblock \href {https://doi.org/10.18653/v1/2021.acl-short.60} {In factuality: Efficient integration of relevant facts for visual question answering}.
\newblock In \emph{Proceedings of the 59th Annual Meeting of the Association for Computational Linguistics and the 11th International Joint Conference on Natural Language Processing (Volume 2: Short Papers)}, pages 468--475, Online. Association for Computational Linguistics.

\bibitem[{Vrande\v{c}i\'{c} and Kr\"{o}tzsch(2014)}]{10.1145/2629489}
Denny Vrande\v{c}i\'{c} and Markus Kr\"{o}tzsch. 2014.
\newblock \href {https://doi.org/10.1145/2629489} {Wikidata: A free collaborative knowledgebase}.
\newblock \emph{Commun. ACM}, 57(10):78–85.

\bibitem[{Wang et~al.(2018)Wang, Wu, Shen, Dick, and van~den Hengel}]{10.1109/TPAMI.2017.2754246}
Peng Wang, Qi~Wu, Chunhua Shen, Anthony Dick, and Anton van~den Hengel. 2018.
\newblock \href {https://doi.org/10.1109/TPAMI.2017.2754246} {Fvqa: Fact-based visual question answering}.
\newblock \emph{IEEE Trans. Pattern Anal. Mach. Intell.}, 40(10):2413–2427.

\bibitem[{Wang et~al.(2017)Wang, Wu, Shen, van~den Hengel, and Dick}]{wang2017fvqa}
Peng Wang, Qi~Wu, Chunhua Shen, Anton van~den Hengel, and Anthony Dick. 2017.
\newblock \href {http://arxiv.org/abs/1606.05433} {Fvqa: Fact-based visual question answering}.

\bibitem[{Wang et~al.(2022)Wang, Yang, Men, Lin, Bai, Li, Ma, Zhou, Zhou, and Yang}]{wang2022ofa}
Peng Wang, An~Yang, Rui Men, Junyang Lin, Shuai Bai, Zhikang Li, Jianxin Ma, Chang Zhou, Jingren Zhou, and Hongxia Yang. 2022.
\newblock \href {http://arxiv.org/abs/2202.03052} {Ofa: Unifying architectures, tasks, and modalities through a simple sequence-to-sequence learning framework}.

\bibitem[{Wang et~al.(2014)Wang, Zhang, Feng, and Chen}]{wang-etal-2014-knowledge}
Zhen Wang, Jianwen Zhang, Jianlin Feng, and Zheng Chen. 2014.
\newblock \href {https://doi.org/10.3115/v1/D14-1167} {Knowledge graph and text jointly embedding}.
\newblock In \emph{Proceedings of the 2014 Conference on Empirical Methods in Natural Language Processing ({EMNLP})}, pages 1591--1601, Doha, Qatar. Association for Computational Linguistics.

\bibitem[{Wu et~al.(2019)Wu, Kirillov, Massa, Lo, and Girshick}]{wu2019detectron2}
Yuxin Wu, Alexander Kirillov, Francisco Massa, Wan-Yen Lo, and Ross Girshick. 2019.
\newblock Detectron2.
\newblock \url{https://github.com/facebookresearch/detectron2}.

\end{thebibliography}

\newpage
\appendix
\begin{table*}[t]
    \centering
    \tiny
    \renewcommand{\arraystretch}{1.15}
    \resizebox{1.8\columnwidth}{!}{%
    \begin{tabular}{lccccc}
    \toprule
    &&&\multicolumn{2}{c}{\textbf{OFA(Ours)}}\\
    \textbf{Types of Questions} & \textbf{MEMNET} & \textbf{UNITIER} & \textbf{Single-Hop} & \textbf{Multi-Hop} \\
    \midrule
    1-Hop & 61.00\% & 65.70\% & 84.25\% & \underline{86.04\%} \\ 
    1-Hop Counting & - & 78.0\% & \underline{88.80\%} & \textbf{90.74\%}\\
    1-Hop Subtraction & - & 28.60\% & \underline{31.25\%} & \textbf{37.89\%}\\
    Multi-Hop & 53.20\% & \underline{87.90\%} & 60.80\% & \textbf{90.40\%}\\
    Boolean & 75.10\% & 94.60\% & \underline{96.89\%} & \textbf{97.17\%}\\
    Comparison & 50.50\% & \underline{90.40\%} & \textbf{90.82\%} & 90.15\% \\
    Counting & 49.50\% & 79.40\% & \underline{90.08\%} & \textbf{90.32\%} \\
    Intersection & 72.50\% & 79.40\% & \underline{87.07\%} & \textbf{89.03\%}\\
    Multi-Entity & 43.50\% & 77.10\% & 84.01\% & \underline{88.53\%} \\
    Multi-Relation & 45.20\% & 75.20\% & 90.10\% & \underline{90.77\%} \\
    Spatial & 48.10\% & 21.20\% & \underline{92.70\%} & \textbf{94.50\%} \\
    Subtraction & \textbf{40.50\%} & 34.40\% & 32.50\% & \underline{40.20\%} \\
    \bottomrule
    \end{tabular}%
    }
    \caption{The table displays the results of all 13 classes on the KVQA dataset. These scores are obtained in a setting where triples are filtered based on both the questions and the images, and the number of triples varies according to a similarity threshold.}
    \label{tab:all13classes}
\end{table*}%

\section{Training Details}
\label{appendix:traindetails}
The experiments encompassed both OFA Base and Large models, maintaining image resolutions at $480 \times 480$ and $640 \times 640$ for Base and Large models, respectively. The dropout rate was set at $0.1$. Adam Optimizer was employed with beta values of $0.9$ and $0.999$, epsilon set to $1 \times 10^{-08}$, and a warm-up ratio of $0.06$. An initial learning rate of $1 \times 10^{-5}$ with polynomial decay was utilized. During test inference, a beam size of $10$ and a temperature of $0.98$ were applied. T5-Base~\citep{raffel2023exploring} model generated embeddings for questions and triples in the triple filtering process. The training was conducted on Nvidia RTX A6000~\footnote{https://www.nvidia.com/en-in/design-visualization/rtx-a6000/}, with each iteration taking approximately $8$ and $12$ hours for the Base and Large models, respectively.\\
We utilized the LLAVA model for fine-tuning the KVQA dataset for the KB-VQA task. Fine-tuning was conducted with a learning rate of $2 \times 10^{-5}$ over $2$ epochs, with a warmup ratio set to $0.03$. All experiments were performed on three Nvidia RTX A6000 GPUs, with each iteration requiring approximately $18$ hours to obtain conclusive results.\\
The number of parameters used and the number of encoder-decoder layers for both the OFA-Base and OFA-Large models are given in Table~\ref{tab:modparameters}.
\begin{table}[h]
\centering
\small
\renewcommand{\arraystretch}{1.25}
\begin{tabular}{|p{1.3cm}|p{1.2cm}|p{1.6cm}|p{1.6cm}|}
\hline
\textbf{Model} & \textbf{\#Param} & \textbf{\#Enc.Layers} & \textbf{\#Dec.Layers} \\ \hline
OFA-Base & 182M & 6 & 6 \\ \hline
OFA-Large & 472M & 12 & 12 \\ \hline
\end{tabular}
\caption{The table displays information regarding the parameter count, as well as the number of encoder and decoder layers for both the OFA Base and OFA Large models.}
\label{tab:modparameters}
\end{table}

\section{Additional Results for KVQA Dataset}
\label{sec:appendixaddkvqa}
In Section~\ref{sec:expres}, we demonstrated the results for 9 classes on the KVQA dataset, aligning with the prior state-of-the-art model, POP-VQA~\citep{Sahu_2024_WACV}. However, in this Section, we extend our analysis to cover all 13 classes within the KVQA dataset, as detailed in Table ~\ref{tab:all13classes}. Additionally, we include the results obtained from the MEMNET~\citep{tai2017memnet} and UNITIER~\citep{chen2020uniter} models for a fair comparison.

We also present results for the other two scenarios. First, no triples are given as context, second when we include all the triples associated with the image, without any filtering based on the question as explained in Section~\ref{appendix:traindetails}. The results for these approaches can be found in Table~\ref{sec:tableforwoin}.
\begin{table}[h]
    \centering
    \tiny
    \renewcommand{\arraystretch}{1.15}
    \resizebox{0.95\columnwidth}{!}{%
    \begin{tabular}{lccc}
    \toprule
    &\multicolumn{2}{c}{\textbf{OFA(Ours)}}\\
    \textbf{Types of Questions} &
    \textbf{With No Triples} & \textbf{With All Triples} \\
    \midrule
    1-Hop & 72.20\% & 76.81\% \\
    1-Hop Counting & 75.95\% & 76.00\% \\
    1-Hop Subtraction & 29.80\% & 30.06\% \\
    Boolean & 86.10\% & 94.40\% \\
    Comparison & 83.59\% & 88.77\% \\
    Counting & 81.10\% & 81.30\% \\
    Intersection & 78.19\% & 76.40\% \\
    Multi-Entity & 71.10\% & 76.32\% \\
    Multi-Hop & 74.22\% & 81.70\% \\
    Multi-Relation & 72.12\% & 83.92\% \\
    Spatial & 89.02\% & 83.43\% \\
    Subtraction & 4.50\% & 7.20\% \\
    \bottomrule
    \end{tabular}%
    }
    \caption{The table presents the performance of various question types in two distinct scenarios: one without the inclusion of any triples as context (referred to as "With No Triples"), and the other with all the relevant triples filtered by images, while not applying any filtering on the questions (referred to as "With All Triples").}
    \label{sec:tableforwoin}
\end{table}%

In our observations, it becomes evident that including all triples results increase in accuracy across most categories when compared to not including any triples at all. However, in more complex categories such as subtraction, the accuracy improvement is not as significant, mainly because accurate predictions demand more refined triples.

An interesting observation occurs when we look at the spatial category. When we provide all triples, accuracy decreases, indicating that in the spatial category, the inclusion of triples is unnecessary. This result shows that our dynamic triple extraction module works effectively, especially in spatial questions, where it rarely provides external triples. This emphasizes that the module can smartly adjust to meet the specific needs of each question.\\
We discuss this in Section~\ref{kvqaresults}
\section{Additional Results for FVQA Dataset}\label{additresFVQA}
In Section ~\ref{sec:fvqa}, due to the absence of labels in the dataset, we utilized the CLIP model to extract relevant triples from the image. To achieve this, we divided the image into four patches, computing the most relevant triples for each patch.
In this Section, we present results for two additional settings to ensure transparency,

\textbf{Triples relevant to Full Image:} In this configuration, we refrain from dividing the image into patches. Instead, we compute relevant triples for the entire image. These results are summarized in Table ~\ref{tab:fullimage}.
\begin{table}[h]
\centering
\small
\renewcommand{\arraystretch}{1.25}
\resizebox{0.46\textwidth}{!}{
\begin{tabular}{|l|l|l|}
\hline
\textbf{Model} & \textbf{Without-fine-tuning} & \textbf{With-fine-tuning} \\ \hline
OFA-Base & 33.28 & 39.94 \\ \hline
OFA-Large & 34.84 & 43.20 \\ \hline
\end{tabular}}
\caption{Results on FVQA dataset. Exact match score with and without fine-tuning on the FVQA dataset. Triples relevant to images are computed by considering the whole image without dividing it into patches.}
\label{tab:fullimage}
\end{table}
The problem with this approach is that when computing the cosine similarity of the CLIP embedding of the entire image and triples, triples relevant to smaller objects might not be captured. For instance, as depicted in Fig~\ref{fig:splitimage}, triples related to the flower vase could be overlooked.

\textbf{Triples relevant to objects in the image:} For extracting the triples relevant to objects in the image we use the following approach:
\begin{itemize}
    \item Bounding Box Extraction: We identify bounding boxes for each object present in the image. These bounding boxes define the spatial regions corresponding to the objects.
    \item Detectron Model: To achieve this, we utilize the Detectron model, which detects the precise coordinates of the bounding boxes.
    \item Image Patch Extraction: Once we have the bounding box coordinates, we extract image patches corresponding to those regions.
    \item Triple Extraction: For each image patch, we find the relevant triples associated with the objects within that patch.
\end{itemize}

The results are demonstrated in table~\ref{tab:objectstable}.
\begin{table}[h]
\centering
\small
\renewcommand{\arraystretch}{1.25}
\resizebox{0.46\textwidth}{!}{
\begin{tabular}{|l|l|l|}
\hline
\textbf{Model} & \textbf{Without-fine-tuning} & \textbf{With-fine-tuning} \\ \hline
OFA-Base & 33.75 & 44.62 \\ \hline
OFA-Large & 38.42 & 46.71 \\ \hline
\end{tabular}}
\caption{Results on FVQA dataset. Exact match score with and without fine-tuning on the FVQA dataset. Triples relevant to images are computed by considering each object in the image.}
\label{tab:objectstable}
\end{table}

In the above two approaches, we filtered the triples based on the image, for further filtering based on a question we used the same method as explained in Section~\ref{dynamictriples}.\\
For prediction we employed a pre-trained model on the KVQA dataset, specifically focusing on the best setting where the model was trained with multi-hop dynamic triples as context. The fine-tuning and inference process also considers a dynamic number of triples as context. We have provided results for both scenarios: without and with fine-tuning on the FVQA dataset, as elaborated in Section~\ref{sec:fvqa}.
\section{Training CLIP Model}\label{CLIPfine-tuning}
CLIP model~\cite{radford2021learning} is trained for image-text similarity and not for image-triples similarity. Therefore, we train the CLIP model to extract triples that are relevant to the image. We denote the set of triples from the knowledge graph as $t_k$, and the reference image as \textit{I}. To identify the triples that are relevant to the reference image, we minimise the following objective, \begin{gather*}
\Large{-\log\frac{\exp(s(I,t_k^{(+)})e^{\tau})}{\exp(s(I,t_k^{(+)})e^{\tau})+\sum_{j}\exp(s(I,t_k^{(j)})e^{\tau})}}
\end{gather*}\\
We implement $s(I,t_k^{(+)})$ using CLIP as:\\
$s(I,t_k^{(+)})=\cos({CLIP_{V}(I),CLIP_{T}(t_{k})})$\\
Here $t_k^{(+)}$ denotes the triple relevant to the image, $t_k^{(j)}$ denotes the irrelevant triples for an image and $\tau$ denotes  temperature parameter
which controls the range of the logits in the softmax as explained in ~\cite{radford2021learning}.
Since there isn’t a specific dataset available for images and their relevant triples, we utilize the ViQuae Wikipedia Corpus~\citep{10.1145/3477495.3531753} to acquire the images and their corresponding triples. We have chosen 2000 instances that include images and their related triples, which were extracted using the Wikidata knowledge graph~\cite{10.1145/2629489}. We train the CLIP model using the above objective to get relevant triples.\\
We discuss this in Section~\ref{sec:fvqa}
\section{Algorithm}\label{algo1}
The algorithm is discussed in ~\ref{exp:algo}
\begin{algorithm}
\footnotesize 
\caption{Retrieving context for k-hop Question Answering and feeding the Question, Image, and Context into a Transformer Encoder-Decoder model to predict the desired answer.}
\label{exp:algo}
\begin{algorithmic}[1]
\Require
\State $Q_0$ $\rightarrow$ Input Question
\State $T$ $\rightarrow$ Triples from Knowledge Graph
\State $k$ $\rightarrow$ Number of Hops
\State $I$ $\rightarrow$ Image
\State $E$ $\rightarrow$ Named Entities
\Ensure
\State \textbf{Triple Filtering (By Images)}
\For{${Count}$ in $k$}
    \For {$(Head, Relation, Tail)$ in Knowledge Graph}
       \If {Head or Tail in $E$}
          \State Relevant Triples += $(Head, Relation, Tail)$
       \EndIf
    \EndFor
\EndFor

\State \textbf{Triple Filtering}

\For {Triple in Relevant Triples}
   \State $T\_Embed$ = T5 Base(Triple)
   \State $Q\_Embed$ = T5 Base($Q_0$)
   \If {{Similarity}(T\_Embed, Q\_Embed) $\geq$ $\lambda$}
      \State Context += Triple
   \EndIf
\EndFor

\State \textbf{Prediction Module}

\State Answer = OFA\_Model($Image\textless SEP\textgreater Question\textless SEP\textgreater Named Entities\textless SEP\textgreater Context$)

\end{algorithmic}
\end{algorithm}
\section{Prompting on LLAVA Model}\label{appendix:llavaprompt}
In this segment, we'll furnish the prompt utilized to find responses from the llava-hf/llava-v1.6-mistral-7b-hf model for image-related questions. To ensure fair comparison based on exact match scores, we want concise answers to avoid any extraneous information. The provided prompt generates concise responses, minimizing any potential noise.\\
\textbf{Prompt for evaluation without giving any knowledge}\\
Please answer concisely in one or two words:\\
Question: <question>\\
Named Entities: <named entities>\\
\textbf{Prompt for evaluation when giving knowledge}\\
Please answer the question concisely in one or two words. We also provide Named Entities and knowledge triples separated by <sep> token for your assistance:\\
Question: <question>\\
Named Entities: <named entities>\\
Triples: <triples string>\\
We discuss this in Section~\ref{prompting}
\section{Baselines on CRIC-VQA dataset}\label{appendix:cricbaselines}
In this section, we explain each baseline in brief as depicted in Table~\ref{tab:cricvqatable}.\\
\textbf{Q-Only GRU} -  Q-Only model only takes the GRU question features as input.\\
\textbf{Q-Only BERT} -  Q-Only model only takes the BERT question features as input.\\
\textbf{SF} - SF first uses visual concepts extracted by object, scene, action predictors, CNN image feature, and LSTM question feature to retrieve the Top-1 related knowledge item, then uses the question feature and retrieved knowledge item to predict the answer.\\
\textbf{Bottom-Up+latt} - Bottom-Up is a traditional VQA model emphasizing object-level reasoning with soft attention to object regions. This baseline enhances Bottom-Up by incorporating a binary cross-entropy loss on attention scores to guide the model to focus on the correct region when combining attended image and question features for generating the final answer.\\
\textbf{MAC-CS} -  MAC is a leading modular VQA model designed for CLEVR and GQA. It breaks down questions into attention-based reasoning steps. The expanded MAC's capabilities to incorporate access to knowledge items resulted in MAC-CS, which focuses on commonsense reasoning.\\
\textbf{NMN-CS} - The Neural Modular Network (NMN) is a distinct VQA model. However, its original iterations are not directly applicable to commonsense questions. To address this limitation, visual commonsense reasoning modules have been integrated, resulting in NMN-CS.\\
\textbf{Memory-VQA+latt} - This memory network operates by encoding input materials such as knowledge items and the image in the CRIC as memories. It utilizes the question to initiate an iterative attention process, enabling the model to retrieve relevant information for answering the question. In contrast to Memory-VQA, this baseline further incorporates a cross-entropy loss on attention scores.\\
\textbf{VILBERT+ERNIE+latt} - The model consists of three modules: ViLBERT for image and question feature extraction, ERNIE for candidate knowledge item feature extraction, and an attention module for predicting the answer by using pooled features from both transformers to locate the target image region.\\
We discuss this in Section~\ref{cricresults}
\section{Qualitative Analysis of LLAVA Generated Answers}\label{humaneval}
We analyzed LLAVA-generated answers on the KVQA dataset in a zero-shot scenario. The primary objective was to confirm whether the model's incorrect answers were a result of its output or an issue with the exact match metric. To achieve this, we sampled a total of random 200 instances containing questions of all 13 classes where the LLAVA model provided incorrect answers. We determined whether the incorrect answers were due to the model itself or a metric-related problem. This evaluation yielded counts for both scenarios: instances where the model's answers were incorrect and instances where the issue lay with the metric. The results of the evaluation are shown in Table~\ref{tab:humaneval}.\\
\begin{table}[h]
    \centering
    \resizebox{0.80\columnwidth}{!}{%
    \begin{tabular}{lc}
    \toprule
    \textbf{Model mispredictions}&\textbf{Metric problem}\\
    \midrule
    182 & 18\\
    \bottomrule
    \end{tabular}%
    }
    \caption{Human Evaluation results for the LLAVA Model Output}
    \label{tab:humaneval}
\end{table}%
The primary issue arises with spatial inquiries where the correct response is "Person on the Left", or "Person on the Right", or "Person on the Center" yet the model tends to provide named entities of individuals instead.\\
\textbf{For example:}\\
\textbf{Question:} Who among the people in the image lived longest?\\
\textbf{Truth Answer:} Person in the left\\
\textbf{Predicted Answer:} Lili Damita\\
For such questions, given the limited options of only three potential answers, we adjust the prompt as:\\
\textbf{Updated prompt for Spatial Class:}\\
Please answer concisely in one or two words:\\
Question: <question>\\
Named Entities: <named entities>\\
Don’t give named entities in the answer instead provide the answer in form Person in Center, Person in Left, Person in Right.\\
The outcomes presented in Table~\ref{tab:llavawwoknow} account for this scenario to guarantee a fair assessment process.\\
Regarding the CRIC-VQA dataset, which focuses on objects and typically elicits responses of one or two words like "Desk", "Water" etc there is no issue with metric and the model generates wrong answers.\\
We discuss this in Section~\ref{cricresults}
\section{OFA Model}\label{ofamodel}
We leverage the power of Unified Vision-Language (VL) modelling~\citep{wang2022ofa}, which has demonstrated significant potential across various VL tasks. 
For our VQA tasks, we adopt a vision language transformer encoder-decoder model OFA Base and OFA Large architecture. The OFA model is designed to handle diverse tasks and modalities, seamlessly integrating vision-only, language-only, and vision-language tasks within a sequence-to-sequence learning framework.

Our input comprises ResNet152~\citep{he2015deep} features extracted from the image, followed by the question and context, both tokenized using byte-pair encoding (BPE)~\citep{bostrom2020byte}. We employ a unified vocabulary that encompasses tokens from both visual and linguistic domains. Transformers serve as the core encoders and decoders, treating the vision-language task as a sequence-to-sequence problem.
\section{More Examples}\label{appendix:moreexamples}
Refer to Table~\ref{my-labellasttable} for the examples used in Section~\ref{sec:quali}, as well as some additional examples that demonstrate the effectiveness of our approach. Table~\ref{my-labellasttable} includes certain questions that do not necessitate any knowledge (as seen in Row 7). These can be addressed solely based on image features, without the need for external knowledge. Supplying triples in these instances results in incorrect predictions. These questions predominantly belong to the spatial category. Additionally, some questions are straightforward and do not require knowledge filtering (as seen in Row 10). Providing all triples without filtering based on questions in these cases would also yield correct answers, eliminating the need for filtering. These questions are primarily 1-hop questions. However, for complex categories such as 1-hop subtraction, multi-hop, etc., a robust reasoning capability is required. Therefore, supplying filtered knowledge is essential to prevent any confusion that could lead to incorrect predictions.\\
We discuss this in Section~\ref{sec:quali}
\begin{table*}[t]
\centering
\setlength{\extrarowheight}{0.5em}
\begin{tabular}{|>{\small}m{2.3cm}|>{\small}m{2.3cm}|>{\small}m{2cm}|>{\small}m{2cm}|>{\small}m{2.5cm}|>{\small}m{1.8cm}|}
\hline
\textbf{Question} & \textbf{True Answer} & \textbf{No Triples} & \textbf{All Triples} & \textbf{Filtered Triples} & \textbf{Image} \\
\hline 
Is the person in the image a politician? & No & Yes & No & No & \includegraphics[scale=0.15,valign=c]{Image6QualitativeAnalysis.jpg} \\[1em] \hline
In which country was the person in the image born? & Slovakia & Hungary & Slovakia & Slovakia & \includegraphics[scale=0.1,valign=c]{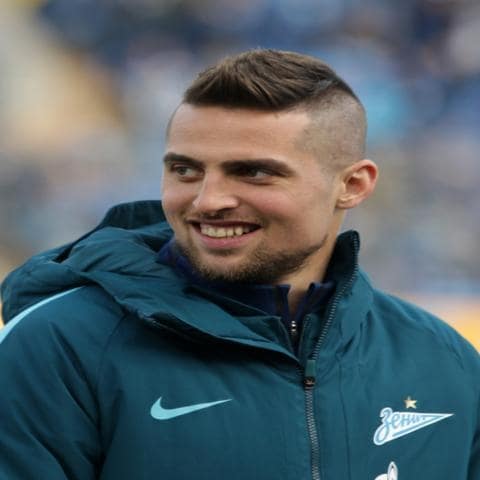} \\[1em] \hline
For how many years did the person in the image live? & 83 & 72 & 82 & 83 & \includegraphics[scale=0.15,valign=c]{Image5QualitativeAnalysis.jpg} \\[1em] \hline
Were all the people in the image born in the same country? & No & Yes & Yes & No & \includegraphics[scale=0.15,valign=c]{Image3QualitativeAnalysis.jpg} \\[1em] \hline
Who among the people in the image ever married Vladimir Soshalsky? & Person on the left & Person on the right & Person on the right & Person on the right & \includegraphics[scale=0.15,valign=c]{Image4QualitativeAnalysis.jpg} \\[1em] \hline
For how many years did the person in the image live? & 79 & 86 & 85 & 79 & \includegraphics[scale=0.1,valign=c]{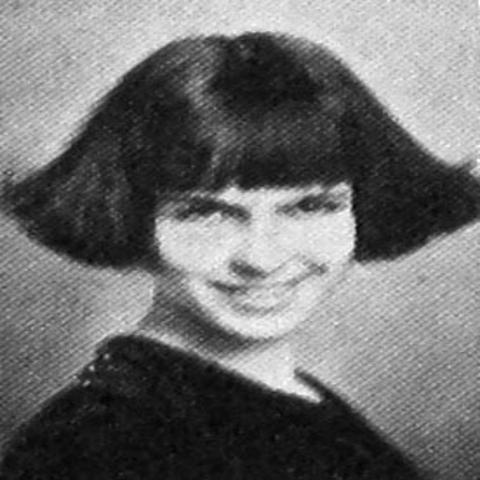} \\[1em] \hline
Do all the people in the image have a common occupation? & No & Yes & Yes & No & \includegraphics[scale=0.1,valign=c]{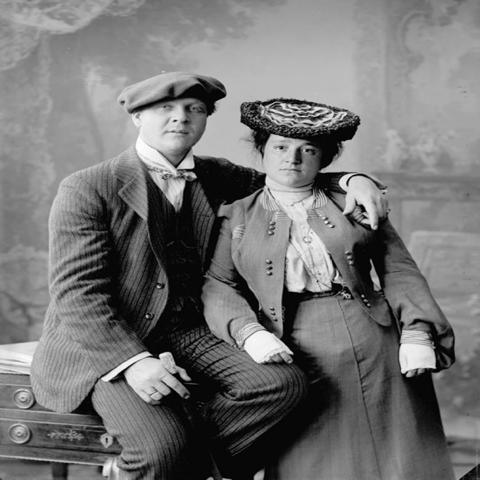} \\[1em] \hline
Who is to the right of Jorge Toriello Garrido? & Jacobo Árbenz &  Jacobo Árbenz & jajaxedlol & Jacobo Árbenz & \includegraphics[scale=0.1,valign=c]{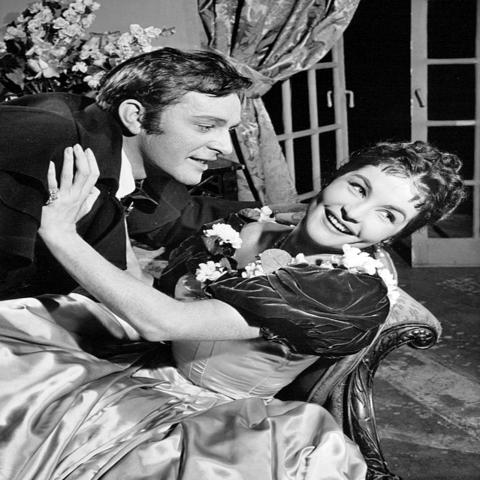} \\[1em] \hline
In which year did the person in the image start professional activities? & 1911 &  1920 & 1986 & 1956 & \includegraphics[scale=0.1,valign=c]{43155.jpg} \\[1em] \hline
Who among the people in the image ever married to Bill Williams? & Person in the right &  Person in the left & Person in the right & Person in the right & \includegraphics[scale=0.08,valign=c]{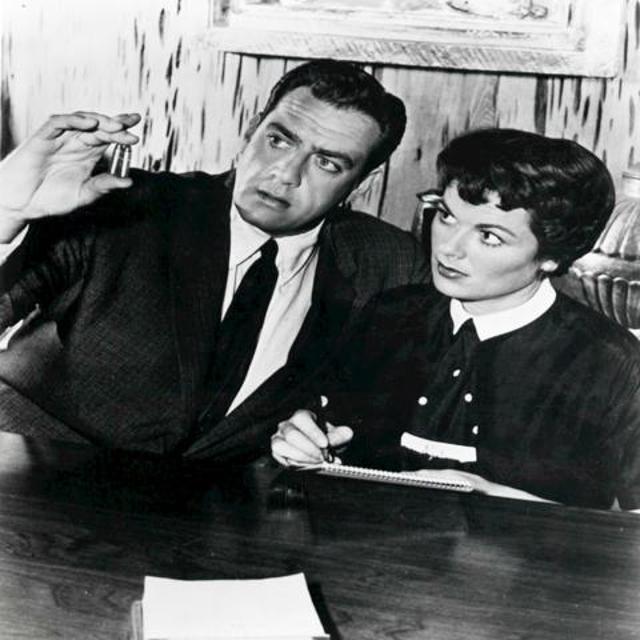} \\[1em] \hline
\end{tabular}
\caption{Error analysis table, presents instances from the datasets and their predicted answers in three settings mainly no triples, all triples and filtered triples.}
\label{my-labellasttable}
\end{table*}

\section{Future Work}\label{appendix:future}
Several potential avenues for future exploration are available. Presently, the fact retriever and answer prediction module undergo separate training processes. Exploring an end-to-end trainable model that seamlessly integrates both components represents an intriguing direction to explore. The optimal number of triplets for context was determined through experimentation, incorporating heuristics for similarity values, among other factors. However, enhancing performance can be achieved through the model's automatic learning of this ideal number of triplets based on the characteristics of the question, image, etc. Exploring additional techniques to enhance the model's generalization across different domains is another compelling direction to investigate. Creating an explanatory model for the retrieved context would prove beneficial for numerous practical applications. We anticipate that the numerous avenues for future work, along with our presented results, will inspire further exploration and advancements in the KBVQA domain.

\label{sec:appendix}

\end{document}